\documentclass[conference]{IEEEtran}
\IEEEoverridecommandlockouts
\usepackage{cite}
\usepackage{amsmath,amssymb,amsfonts}
\usepackage{algorithmic}
\usepackage{graphicx}
\usepackage{textcomp}
\usepackage{xcolor}
\usepackage{silence}
\def\BibTeX{{\rm B\kern-.05em{\sc i\kern-.025em b}\kern-.08em
    T\kern-.1667em\lower.7ex\hbox{E}\kern-.125emX}}
\begin{document}

\title{Semantic Driven Energy based Out-of-Distribution Detection\\
}

\author{
\IEEEauthorblockN{Abhishek Joshi}
\IEEEauthorblockA{\textit{Samsung R\&D Institute} \\
Bangalore, India \\
abhi.joshi@samsung.com
\vspace{-4mm}
}
\and
\IEEEauthorblockN{Sathish Chalasani}
\IEEEauthorblockA{\textit{Samsung R\&D Institute} \\
Bangalore, India \\
sathish.c@samsung.com
\vspace{-4mm}
}
\and
\IEEEauthorblockN{Kiran Nanjunda Iyer}
\IEEEauthorblockA{\textit{Samsung R\&D Institute} \\
Bangalore, India \\
kiran.iyer@samsung.com
\vspace{-4mm}
}
}

\maketitle

\begin{abstract}
Detecting Out-of-Distribution (OOD) samples in real world visual applications like classification or object detection has become a necessary precondition in today’s deployment of Deep Learning systems. Many techniques have been proposed, of which Energy based OOD methods have proved to be promising and achieved impressive performance. We propose semantic driven energy based method, which is an end-to-end trainable system and easy to optimize. We distinguish in-distribution samples from out-distribution samples with an energy score coupled with a representation score. We achieve it by minimizing the energy for in-distribution samples and simultaneously learn respective class representations that are closer and maximizing energy for out-distribution samples and pushing their representation further out from known class representation. Moreover, we propose a novel loss function which we call Cluster Focal Loss(CFL) that proved to be simple yet very effective in learning better class wise cluster center representations. We find that, our novel approach enhances outlier detection and achieve state-of–the-art as an energy-based model on common benchmarks. On CIFAR-10 and CIFAR-100 trained WideResNet, our model significantly reduces the relative average False Positive Rate(at True Positive Rate of 95\%) by 67.2\% and 57.4\% respectively, compared to the existing energy based approaches. Further, we extend our framework for object detection and achieve improved performance.
\end{abstract}


\section{Introduction}
Deploying reliable machine learning systems in safety-critical applications like biometric authentication, medical diagnosis or autonomous driving is of paramount importance. Not only safety critical but classification and object detection solutions deployed to mobile, e-commerce applications require a robust model for best user experience. The inductive bias for the above mentioned applications is generally very high with models trained through supervised learning, as we violate the most basic i.i.d (independent and identically distributed) assumption, that assumes that training data and real world data we encounter during inference are independent and identically distributed. In reality, these applications are subjected to deal with data that belongs to different distributions altogether. Modern neural networks are most vulnerable when trained on particular data distribution and inferred on samples belonging to a distribution far from training distribution (called out-of-distribution (OOD) samples or outliers). This vulnerability motivates us in designing more robust and foolproof systems for OOD detection.

\begin{figure}[th!]
  \includegraphics[width=0.99\linewidth, height=16.3cm]{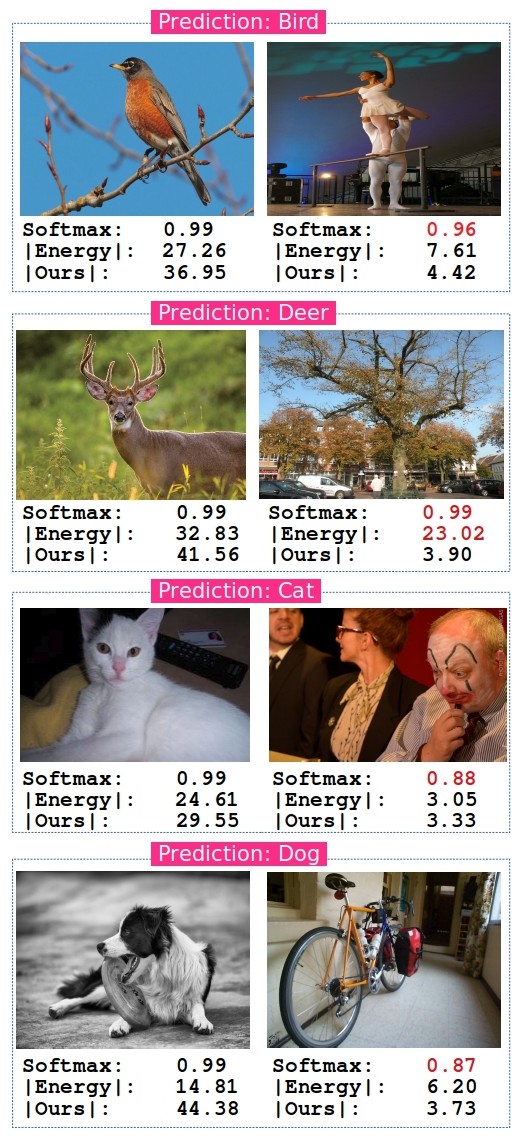}
   \caption{Softmax based classifiers perform well on IN-distribution data, however, fires  false predictions(scores in red colour) for OOD data samples with high confidence. Our proposed framework(CFL-MLSE) scores are much more reliable and robust against such distribution, compared to softmax and energy\cite{liu2020energy} model.}
\label{fig:FP_Predictions}
\vspace{-5mm}
\end{figure}

Supervised learning approaches produce semantic representations that can discriminate classes labeled in the training dataset, relying on softmax confidence. However, softmax based OOD detection approaches fail often as they can produce high confidence scores even for OOD samples. To overcome that, recently \cite{liu2020energy} proposed an energy based training method to output energy score to detect OOD samples. Although effective in discriminating OOD samples, it lacks to impart better discriminative representation to establish large margin between in-distribution samples and out-distribution samples.

\hyphenpenalty=100000 Many approaches \cite{hassen2020learning} have been proposed to improve discriminative power of learned features. We take our motivation from linear discriminant analysis and K-means clustering, and propose a framework for OOD detection which is two-fold:
\begin{itemize}
  \item \hyphenpenalty=100000 We minimize the intra-class variations to have compact class cluster representation while keeping outliers separated from all class clusters, to learn representations that enhance its discriminative power to detect outliers. Further, we propose a novel loss function called cluster focal loss that can enhance the representations of class wise cluster centers with maximum inter class separation.
  \item We couple metric learning based distance function with the energy function to jointly minimize the score for inliers and maximize the score for outliers, the learned score separates inliers from outliers during inference. We call the joint score as semantic energy score (SE score) and propose several variants of the framework.
\end{itemize}
 Our approach exceeds state-of-the-art on OOD test sets. At the same time, the method enhances accuracy for in-distribution test. As shown in Figure \ref{fig:FP_Predictions}, when a trained model is subjected to open world images, more often than not, we cannot entirely rely on softmax confidence alone. It can be seen that in certain scenarios modeling energy alone would not suffice due to visual similarities in OOD samples compared to samples in in-distribution. These are the tough cases that can be resolved through our methodology by bringing semantic information to model energy.

\section{Related Work}
In machine learning, the techniques of Openset Recognition (OSR) and Out-of-Distribution detection have very subtle differences between them. Sometimes the terms are used synonymously in literature. 

Strictly speaking, the goal of open set recognition is to accurately classify new and unknown data that belongs to training distribution and reject data that does not belong to this distribution. OOD methods on the other hand models to determine if an input data sample belongs to training distribution and not concerned about correct classification, if data sample belongs to in-distribution. Despite differences in approaches and subtleties in the techniques, we emphasize a hybrid approach that can complement the short falls in each method.

Energy based models have a long history in the fields of physics, statistics and machine learning. \cite{lecun2006tutorial, song2021train} have shown that Energy Based Models (EBM) rather than being specified as normalized probability, they can be specified as negative log-likelihood probability. In doing so, one doesn't have to calculate normalizing constant, also called partition function, which is intractable more often. With this EBMs have found wide applications in many fields of machine learning like density estimation \cite{wenliang2019learning, song2020sliced} for statistically modeling to fit data, discriminative learning \cite{grathwohl2019your, gustafsson2020energy} for classification and regression, reinforcement learning \cite{haarnoja2017reinforcement} for learning energy based policies for continuous states and actions, natural language processing \cite{mikolov2013distributed, deng2020residual} for learning syntactic and semantic distributed vector representations and generative modeling \cite{ngiam2011learning, du2019implicit, xie2016theory} for image generation.

\subsection{Open-Set Recognition}
We have various OSR methods in literature that employ different data strategies to perform open set recognition. \cite{neal2018open} tries to generate examples through GANs that are visually close to training examples and yet do not belong to any training category.  Some methods use unknown data to learn characteristics that separate from known distribution. \cite{ditria2020opengan} uses a conditional GAN based method conditioned on feature embedding drawn from a metric space to generate samples belonging to out-of-distribution novel classes. \cite{kong2021opengan} uses GAN to augment open data in two ways, one by generating fake data based on open set samples, second by generating intermediate features for open-set. Both features and images are used to train discriminator. \cite{dhamija2018reducing} designs novel losses to maximize entropy for unknown inputs. They also modify magnitudes of deep feature space to increase separation. \cite{bendale2016towards} modifies the softmax layer of the neural network. The scores in the penultimate layer are redistributed to accommodate for unknown class. Weibull distribution is fit to Mean Activation Vectors(MAV) of each class. During inference, depending on parameters of learned Weibull distribution, scores are redistributed to recognize unknown classes. Few methods do not require additional data. They try to learn the underlying structure of known distribution to distinguish from unknown distribution. \cite{hassen2020learning} introduces inter-intra loss (abbreviated as ii-loss) to bring intra classes together and separate inter classes in their deep feature representation. We borrow inspiration from this method to model in-distribution classes to have better inter class separability in high dimensional feature space through our novel cluster focal loss function. This inter class separation maintain accuracy of inlier samples during inference.

\subsection{Out of Distribution Detection}
OOD detection methods in literature follow several strategies to detect novel or outlier samples. Few are distance based detection methods while some are classification based detection methods. \cite{lee2017training} train classifier to be less confident on unknown distribution at the same time generating training samples similar to unknown distribution samples. This is classification based detection method with GANs. There are also various detection score methods proposed like prediction entropy \cite{mohseni2020self}, KL-Divergence score \cite{hendrycks2019using}. \cite{hsu2020generalized} proposed a generalized Out-of-Distribution Image Detection(ODIN) method to increase the gap in softmax classifier for inlier and outlier samples. Interestingly, Grathwohl \textit{et al.} \cite{grathwohl2019your} has shown that joint energy based model training implicitly improves calibration, robustness and OOD detection. Liu \textit{et al.} \cite{liu2020energy} prove that there exists a direct relation between output of a network after the softmax layer and Gibbs distribution of class specific energy values. We extend this concept to multi-levels in the penultimate layers of the network. In doing so, we induce sparsity of activations in channels that don't fire for a pattern and boost the density of activations in channels that fire for a pattern. To the best of our knowledge, ours is a novel attempt to introduce this concept. This would enhance the separability between inliers and outliers.

\section{Methodology}
\label{section:methodology}
Its critically important to detect outliers in safety critical applications, however, it is also equally important to maintain good classification accuracy for in-distribution data simultaneously. Hence, we propose an end-to-end trainable loss formulation that is based on two objectives: \\
\textit{i.)} Shape the energy surface of the network to separate inliers from outliers through Energy based modeling. \\
\textit{ii.)} Integrate semantics into the energy model. For better semantic representation class wise, we employ clustering in feature space by our novel Cluster Focal Loss.

\begin{figure}[t]
\begin{center}
   \includegraphics[width=0.99\linewidth]{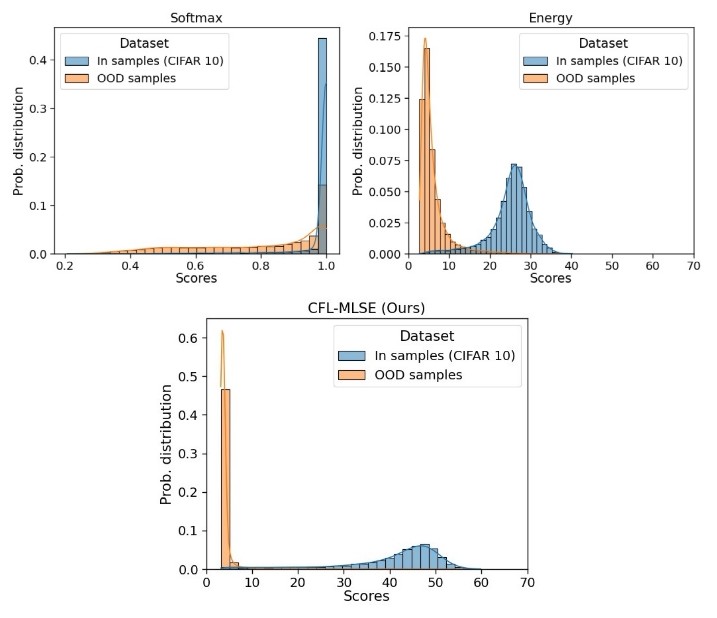}
\end{center} \vspace{-5mm}
   \caption{Score distribution for Softmax, Energy\cite{liu2020energy} and our proposed technique, for in- and out-distribution samples. Our proposed CFL-MLSE method clearly leads to a better energy gap.}
\label{fig:score_dist}
\end{figure}

\subsection{Problem Formulation}

The core of the energy-based model (EBM) \cite{liu2020energy} is to provide a function  $E(x) =\mathbb{R^D} \longrightarrow \mathbb{R}$ that maps each point $x$
of an input space to a single scalar referred to as energy. A collection of energy values could be turned into a probability density $p(x)$ through the Gibbs distribution as follows:

\begin{equation}
\label{eq1}
    p(y | x) = \frac{e^{-E(x,y)/T}}{\int_{y'}e^{-E(x,y')/T} }
\end{equation}

where $T$ is the temperature parameter. The energy-based model has an inherent connection with classification models in modern machine learning. Consider a neural network classifier $f(x): \mathbb{R^D} \longrightarrow \mathbb{R^K}$, which maps an input $x \in \mathbb{R^D}$ to $K$ real-valued numbers known as logits. These logits are used to derive a categorical distribution using the well-known softmax function. As derived in \cite{liu2020energy}, an energy for a given input $(x,y)$ can be defined as $E(x,y) = -f_y(x)$ where $f_y(x)$ indicates the $y^{th}$ index of $f(x)$ i.e., the logit corresponding to the $y^{th}$ class label. The free energy function $E(x;f)$ over $x \in \mathbb{R^D}$ can be expressed in terms of denominator of the softmax activation: 

\begin{equation}
\label{eq2:free}
    E(x;f) = -T.\log \sum_{i}^{k}e^{f_{i}(x)/T }
\end{equation}

Due to limitations in the existing energy framework indicated earlier, we propose and formulate a novel semantic driven energy-based framework that incorporates the semantic cluster distances through cosine similarity into the energy scoring function.

\subsection{Proposed Approach}

\subsubsection{Semantic Driven Energy-bounded Learning}
As proposed in \cite{liu2020energy}, through an energy-bounded learning objective the neural network is fine-tuned to create an energy difference by assigning lower energies to the in-distribution data, and higher energies to the OOD data. Additionally, we propose to couple metric learning based distance function with the energy function to explicitly minimize the joint objective for in-distribution samples and maximize the score for out-distribution samples. We refer this loss function as semantic energy loss and train our energy-based classifier via following objective: 

\begin{equation}
\label{eq:optimization}
    \min_{\theta} \mathbb{E}_{(x,y)\sim\mathcal{D}_{in}^{train}}[-\log F_{y}(x)]  +  \lambda.L_{sem\_energy}
\end{equation}

where $F(x)$ is the softmax output of the classification model and  $\mathcal{D}_{in}^{train}$ is the in-distribution training data. The scalar hyperparameter $\lambda$ is used to weigh the semantic energy loss.

The overall training objective combines the standard cross-entropy loss, along with an energy loss defined in terms of semantic energy: 
\begin{equation}
\begin{aligned}
L_{sem\_energy} = \mathbb{E}_{(x_{in},y) \sim\mathcal{D}_{in}^{train}}(\max(0,E_s(x_{in}) - m_{in}))^2 \\
+ \mathbb{E}_{(x_{out},y) \sim\mathcal{D}_{out}^{train}}(\max(0, m_{out} - E_s(x_{out}) ))^2
\label{eq:semantic_energy}
\end{aligned}
\end{equation}

where $\mathcal{D}_{out}^{train}$ is the unlabelled auxiliary OOD training data. We use squared hinge loss with dual margin hyper-parameters  $m_{in}$ and $m_{out}$ to penalize out of bound positive and negative samples in train data to help model learn better energy gaps.
\begin{equation}
E_s(x;f) = -T.\log \sum_{i}^{k}e^{Z_{i}(x)/T }
\label{eq:sem_score}
\end{equation}

\begin{equation}
Z_i(x) = SIM_{i}(x).f_i(x)
\label{eq:z}
\end{equation}

\begin{equation}
SIM_i(x) = \frac{f(x).M_i}{\|f(x)\| \|M_{i}\|}
\label{eq:similarity}
\end{equation}

where $SIM_i(x)$ is defined as cosine similarity between the logit vector $f(x)$  ($f_{i}(x)$ is logit of $i^{th}$ class) and $M_i$ i.e. class mean activation vector corresponding to the $i^{th}$ class label. 

Comparing Eq. \ref{eq2:free} and Eq. \ref{eq:sem_score}, it is clear that both equations have similar forms and illustrate that our semantic driven formulation fits naturally to an energy based framework.

\textbf{Learning cluster representations:}
Firstly, in practice, the matrix $M$ could be initialized leveraging a pretrained softmax based classifier. The pretrained logits serve as a good prior for mean vector initialization. Other options can also be through cluster based learning approaches like minimizing the ii-loss \cite{hassen2020learning} which encourages separation between classes in a learned representation space. However, for effective cluster representation learning we propose a novel loss function, which we call Cluster Focal Loss(CFL).

 During training, the model is first trained for a few iterations with cross entropy loss to get a good initial estimate of cluster means. Post that, we calculate the class wise cluster means on train data. Once the initial estimate of the cluster center means is estimated, we start training with CFL objective loss and constantly update the cluster means from each mini-batch using exponential moving average(EMA) and store it as part of the model. Further, the versatility of the obtained matrix $M$ capturing the semantic information is not just confined to training, but we propose to leverage it during inference as well, as described in Eq. \ref{eq:semantic_inference_eq}

\textbf{Proposed Cluster Focal loss:}
Our motivation is to get maximum separation between classes by learning a better representation in large dimensional spaces. To this end, we propose a novel loss function. Our proposed Cluster Focal Loss is an intuitive, simple and yet effective loss function that works well with any softmax based learning objective. Our loss objective is inspired from the focal loss \cite{Lin2017FocalLF}. In contrast to improving classification accuracy where Focal Loss is usually applied on, we observe that learning better class wise cluster representation also depends on two key factors. One, the need to mitigate the ill-effects of large class imbalances that are usually encountered during training.  Second, the need to differentiate between hard and easy examples, so we can down-weight easy ones and focus on the hard ones. Our method is simple, we calculate scaled semantic similarity of logits wrt. cluster centres. We define the formulation of our novel loss function as:

\begin{equation}
CFL(S(x)) = -\alpha(1 - S(x))^\gamma\log(S(x)) 
\label{eq:cluster_focal_loss}
\end{equation}

where $\gamma \geq 0$ is the tunable focusing parameter. A weighting factor $\alpha \in$ [0,1] is introduced for each class based on cross-validation or set to a scalar value for simplicity. $S(x)$ is the softmax applied on scaled semantic similarity vector, where for each $i^{th}$ class the similarity value is $SIM_i(x)$ as defined in Eq. \ref{eq:similarity}.
The effectiveness of training with our proposed loss functions is summarized in Table~\ref{table:result}, where the model particularly trained with proposed CFL (refer to our method in Table \ref{table:result}) yields superior results, against prior state-of-the-art methods. To the best of our knowledge, ours is the first attempt to introduce a focal loss based method for learning class wise cluster centers.

\textbf{Multi Layer energy training:} We study the behaviour of energy training on multiple layers simultaneously end-to-end. We chose to include the final three layers of the final resnet block to train with our proposed semantic energy formulation. We run multiple experiments to demonstrate that the energy surface of few final layers of the network can be easily modeled for better OOD detection without loss in in-distribution accuracy. We observe that it becomes harder to model lower layers as it deteriorates accuracy of the model. We propose multiple layer energy training. We employ accumulated multiple layer vanilla energy along with CFL based semantic energy formulation for final layer to build the total energy score. We have benchmarked our CFL loss against other popular cluster methods like ii-loss~\cite{hassen2020learning} (refer to the results in Table \ref{table:ablation}). SE in the table indicates semantic energy based formulation. MLSE indicates Multi-Layer Semantic Energy formulation. SE and MLSE employ ii-loss to learn cluster center representation. CFL-MLSE(our proposed method in Tables \ref{table:result}, \ref{table:results_details_C10}, \ref{table:results_details_C100}) indicates CFL based Multi-Layer semantic energy formulation.

\begin{table}[t]
\begin{center}
\begin{tabular}{l|l|c|c|c}
\hline
\textbf{In-dataset} & \textbf{Method} & \multicolumn{1}{l|}{\textbf{FPR95}} & \multicolumn{1}{l|}{\textbf{AUROC}} & \multicolumn{1}{l}{\textbf{AUPR}} \\ \hline \hline
CIFAR-10            & Softmax         & 51.04                               & 90.90                               & 97.92                              \\
& ODIN \cite{liang2018enhancing}          &        35.71                         &    91.09                            & 97.62                            \\
& Mahalanobis \cite{Lee2018ASU}         &     37.08                            &           93.27                     &       98.49                      \\
& OE \cite{hendrycks2018deep}          &  8.53                               &     98.3                           &     99.63                        \\
                    & Energy \cite{liu2020energy}         & 4.92                                & 98.76                               & 99.72                              \\
                    & Ours & \textbf{1.61} & \textbf{99.51}  & \textbf{99.89} \\
                    \hline
CIFAR-100           & Softmax         & 80.41                               & 75.53                               & 93.93       \\
& ODIN\cite{liang2018enhancing}         &   74.64                              &  77.43                              &  94.23                           \\
& Mahalanobis \cite{Lee2018ASU}         &   54.04                              &  84.12                              &    95.88                         \\ 
& OE \cite{hendrycks2018deep}         &  58.10           &   85.19             & 96.40                               \\
                    & Energy \cite{liu2020energy}         & 29.14                               & 94.32                               & 98.74                              \\
                    & Ours & \textbf{12.41} & \textbf{97.18}  & \textbf{99.37} \\
                    \hline
\end{tabular}
\end{center}
\caption{Comparison of OOD detection methods (averaged over 5 OOD datasets). Bold represents superior results.} \vspace{-6mm}
\label{table:result}
\end{table}

\begin{table}[t]
\begin{center}
\begin{tabular}{|l|c|c|c|c|}
\hline
\textbf{OOD Testset} & Method & \textbf{FPR95} & \textbf{AUROC} & \textbf{AUPR} \\
\hline\hline
& Softmax & 59.28 & 88.5 & 97.16 \\
& ODIN \cite{liang2018enhancing} & 49.12  & 84.97  & 95.28 \\
TEXTURES & Mahalanobis \cite{Lee2018ASU}  & 15.0 & 97.33 & 99.41 \\
& OE \cite{hendrycks2018deep} & 12.94 & 97.73 & 99.52 \\
 & Energy\cite{liu2020energy} & 2.79 & 99.05 & 99.75 \\
& Ours & \textbf{0.67} & \textbf{99.73} & \textbf{99.94} \\

\hline

& Softmax & 48.49 & 91.89 & 98.27 \\
& ODIN \cite{liang2018enhancing} & 33.55 & 91.96 & 98.0 \\
SVHN & Mahalanobis \cite{Lee2018ASU}  & 12.89  & 97.62  & 99.47  \\
& OE \cite{hendrycks2018deep} & 4.36 & 98.63  & 99.74  \\
 & Energy\cite{liu2020energy} & 9.31 & 98.06 & 99.59 \\
& Ours & \textbf{2.23} & \textbf{99.52} & \textbf{99.90} \\
\hline

& Softmax & 59.48  & 88.2  & 97.1 \\
& ODIN \cite{liang2018enhancing} & 57.40 & 84.49 & 95.82 \\
PLACES365 & Mahalanobis \cite{Lee2018ASU}  & 68.57 & 84.61 & 96.2  \\
& OE \cite{hendrycks2018deep} & 19.07 & 96.16  & 99.06  \\
 & Energy\cite{liu2020energy} & 9.07 & 97.87 & 99.51 \\
& Ours & \textbf{3.25} & \textbf{99.15} & \textbf{99.82} \\
\hline

& Softmax & 52.15 & 91.37  & 98.12 \\
& ODIN \cite{liang2018enhancing} & 26.62 & 94.57 & 98.77 \\
LSUN & Mahalanobis \cite{Lee2018ASU}  & 42.62 & 93.23 & 98.6 \\
& OE \cite{hendrycks2018deep} & 5.59 & 98.94 & 99.79 \\
 & Energy\cite{liu2020energy} & 2.54 & 99.22 & 99.83 \\
& Ours & \textbf{1.59} & \textbf{99.37} & \textbf{99.85} \\
\hline

& Softmax & 56.03 & 89.83 & 97.74 \\
& ODIN \cite{liang2018enhancing} & 32.05 & 93.50 & 98.54 \\
iSUN & Mahalanobis \cite{Lee2018ASU}  & 44.18 & 92.66 & 98.45  \\
& OE \cite{hendrycks2018deep} & 6.32 & 98.85 & 99.77  \\
& Energy\cite{liu2020energy} & 0.87 & 99.63 & 99.93 \\
& Ours & \textbf{0.32} & \textbf{99.77} & \textbf{99.95} \\
\hline

\end{tabular}
\end{center}
\caption{Detailed comparison of several OOD detection methods, on individual OOD datasets. WideResNet is trained on CIFAR-10 as in-distribution dataset and tested on standard OOD datasets.}
\label{table:results_details_C10}
\vspace{-5mm}
\end{table}

\subsubsection{At Inference: Semantic Energy score as OOD Score}
\label{section:semantic_energy_inference}
Our proposed semantic energy (SE score) serves as a scoring function that is able to distinguish between in- and out-of-distribution in a more discriminative way compared to the vanilla energy framework \cite{liu2020energy}. Inspired from \cite{liu2020energy}, we propose semantic driven energy-based inference using the function $E_s(x;f)$ in Eq. \ref{eq:sem_score} for OOD detection:

\begin{equation}
G_s(x;\tau,f) = \begin{cases}
                0 &  if - E_s(x;f) \leq \tau \\
                1 &  if - E_s(x;f) > \tau 
                \end{cases}
\label{eq:semantic_inference_eq}
\end{equation}

where $\tau$ is the semantic energy threshold. For benchmarking purposes, we choose the threshold using in-distribution data so that a high fraction of inputs are correctly classified by the OOD detector $G_{s}(x)$ The proposed SE score can be easily calculated via the \textit{logsumexp} operator.

\section{Experiments and Results}
In this section, we benchmark our approach in comparison with state-of-the-art on image classification task. We demonstrate the effectiveness of our approach on a wide range of OOD evaluation benchmarks.

\subsection{Setup}

\textbf{Dataset:} 
We use CIFAR-10 and CIFAR-100 as our in-distribution datasets, ImageNet\footnote{Note that previous work has used 80 Million TinyImages as the outlier dataset for training, which has now been withdrawn from the community. Thus, we have used ImageNet for all our experiments.} as outlier dataset for training. We use the standard split for each dataset. Like the train data setup in prior work \cite{liu2020energy}, we also remove all images from ImageNet that have overlap with CIFAR-10 and CIFAR-100. For instance, there are 61K and 267K images in ImageNet data belonging to categories common in CIFAR-10 and CIFAR-100 respectively, and thus removed from the train set. For OOD testing, we use 5 common OOD datasets: SVHN \cite{Netzer2011ReadingDI}, Places365 \cite{Zhou2018PlacesA1}, Texture\cite{cimpoi2014describing}, LSUN \cite{yu2015lsun} and iSUN \cite{Xu2015TurkerGazeCS} for testing.

\textbf{Evaluation Metrics:} 
We compare our approach with the state-of-the-art approaches on three diverse metrics: (1) \textit{FPR95} - the false positive rate of OOD samples when true positive rate of in-distribution samples is 95\% (lower the better); (2) \textit{AUROC} - area under the receiver operating curve (higher the better); (3) \textit{AUPR} - area under the precision-recall curve (higher the better).

\subsection{Results}

\textbf{Training Details:} For a fair comparison we chose the network architecture as WideResNet \cite{zagoruyko2016wide} architecture with 32x32 resolution as used in previous approaches to train all the image classification models. In our experiments, the weight $\lambda$ of $L_{sem\_energy}$ is 0.1 and temperature parameter $T=1$  In consistent with the training settings as in \cite{liu2020energy}, the batch size is 128 for in-distribution data and 256 for unlabeled OOD training data. In this paper, we use PyTorch\cite{NEURIPS2019_9015} for implementation of our models.

\begin{figure}[t]
\begin{center}
   \includegraphics[width=0.99\linewidth]{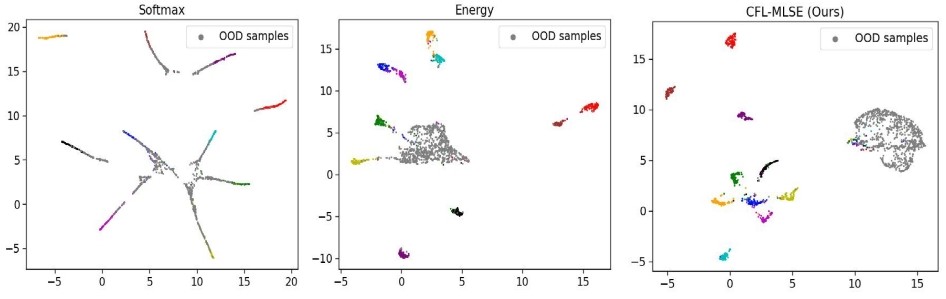}
   \includegraphics[width=0.99\linewidth]{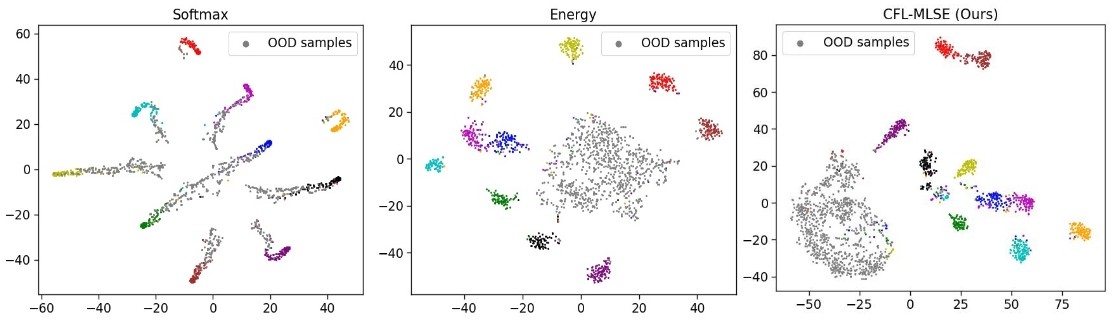}
\end{center}
   \caption{UMAP (first row) and t-SNE (second row)  representation for Softmax, Energy \cite{liu2020energy} and our CFL-MLSE for CIFAR-10 classes and OOD samples.} \vspace{-4mm}
\label{fig:tsne}
\end{figure}

\subsubsection{Qualitative Results}
Firstly, we showcase qualitative results of our approach. In Figure~\ref{fig:score_dist}, we compare the score distribution for in-distribution (CIFAR-10) and out-distribution samples for Softmax approach, vanilla Energy~\cite{liu2020energy} and our proposed CFL based Multi-layer Semantic Energy approach(CFL-MLSE). We observe that Softmax scores are heavily overlapping for in-distribution and out-of-distribution samples, leading to a large number of mis-classifications in OOD samples. Our semantic energy approach significantly reduces the overlap in scores between in- and out-samples as compared to Softmax and vanilla energy formulation. Thus, demonstrating the effectiveness in accurately separating OOD samples from in-distribution samples.

Next, in Figure \ref{fig:tsne}, we compare the 2-dimensional UMAP \cite{UMAP} and t-SNE \cite{maaten2008visualizing} representations of learned features for in- (CIFAR-10) and out-samples from the penultimate layer of WideResNet. We observe that Softmax produces overlapping clusters where OOD samples lie in and around the in-class clusters. On the other hand, our CFL-MLSE approach produces semantic preserving distinctive clusters where OOD samples are far off from the in-class clusters.

\begin{figure}[t]
   \includegraphics[width=0.99\linewidth]{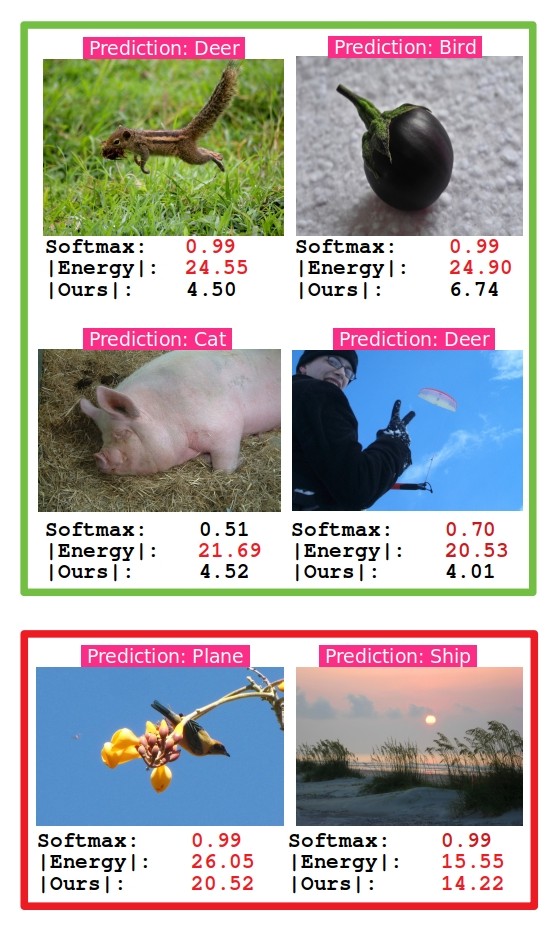} \vspace{-2mm}
   \caption{Green Box contains a set of images where CFL-MLSE predicts lower, and thus better absolute scores for detecting outlier samples compared to Softmax probability and Energy score~\cite{liu2020energy}. Red Box contains a set of images where all the methods failed for OOD-detection. }
\label{fig:FP_Predictions_good_bad}
\end{figure}

Furthermore, we present more subjective results on out-of-distribution data samples. In Figure~\ref{fig:FP_Predictions_good_bad}, for the images enclosed in green coloured box, our proposed CFL-MLSE predicts much lower absolute scores for outlier images. On the other hand, predictions in terms of softmax probability and absolute Energy score~\cite{liu2020energy} are much higher leading to false positives. Thus, CFL-MLSE score serves as a suitable method for OOD detection task. However, the red coloured box represents a set of images for which all the three methods failed in detecting them as out-of-distribution as all the scores are relatively high. For instance, a bird is predicted as a plane. A plausible reason could be attributed to close resemblance of visual cues in the image leading to this confusion.

Figure ~\ref{fig:score_dist} and Figure \ref{fig:tsne} show that our CFL based Multi Layer Semantic Energy approach helps in significantly lowering the OOD mis-classification rate while preserving the semantics of in-distribution classes.

\begin{table}[t]
\begin{center}
\begin{tabular}{|l|c|c|c|c|}
\hline
\textbf{OOD Testset} & Method & \textbf{FPR95} & \textbf{AUROC} & \textbf{AUPR} \\
\hline\hline
& Softmax & 83.29 & 73.34 & 92.89 \\
& ODIN \cite{liang2018enhancing} & 79.27 & 73.45 & 92.75 \\
TEXTURES & Mahalanobis \cite{Lee2018ASU}  & 39.39 & 90.57 & 97.74 \\
 & OE \cite{hendrycks2018deep} & 61.11 & 84.56 & 96.19 \\
 & Energy\cite{liu2020energy} & 4.83  & 98.66  & 99.71  \\
& Ours & \textbf{4.40} & \textbf{98.82} & \textbf{99.75} \\
\hline

& Softmax & 84.49 & 71.44 & 92.93 \\
& ODIN \cite{liang2018enhancing} & 84.66 & 67.26 & 91.38 \\
SVHN & Mahalanobis \cite{Lee2018ASU}  & 57.52 & 86.01 & 96.68 \\
 & OE \cite{hendrycks2018deep} & 65.91 & 86.66 & 97.09 \\
 & Energy\cite{liu2020energy} & 19.81 & 96.33 & 99.23 \\
& Ours & \textbf{7.88} & \textbf{98.09} & \textbf{99.56} \\
\hline

& Softmax & 82.84 & 73.78 & 93.29 \\
& ODIN \cite{liang2018enhancing} & 87.88 & 71.63 & 92.56 \\
PLACES365 & Mahalanobis \cite{Lee2018ASU}  & 88.83 & 67.87 & 90.71 \\
& OE \cite{hendrycks2018deep} & 57.92 & 85.78 & 96.56 \\
 & Energy\cite{liu2020energy} & 12.12 & 97.7 & 99.52 \\
& Ours & \textbf{11.6} & \textbf{97.81} & \textbf{99.54} \\
\hline

& Softmax & 82.42 & 75.38 & 94.06 \\
& ODIN \cite{liang2018enhancing} & 71.96 & 81.82 & 95.65 \\
LSUN & Mahalanobis \cite{Lee2018ASU}  & 21.23 & 96.0 & 99.13 \\
& OE \cite{hendrycks2018deep} & 69.36 & 79.71 & 94.92 \\
& Energy\cite{liu2020energy} & 58.32 & 88.24 & 97.3 \\
& Ours & \textbf{21.4} & \textbf{94.85} & \textbf{98.78} \\
\hline

& Softmax & 82.8 & 75.46 & 94.06\\
& ODIN \cite{liang2018enhancing} & 68.51 & 82.69 & 95.80 \\
iSUN & Mahalanobis \cite{Lee2018ASU}  & 26.10 & 94.58 & 98.72 \\
& OE \cite{hendrycks2018deep} & 72.39 & 78.61 & 94.58 \\
& Energy\cite{liu2020energy} & 50.63 & 70.70 & 97.95 \\
& Ours & \textbf{16.75} & \textbf{96.35} & \textbf{99.22} \\
\hline

\end{tabular}
\end{center}
\caption{Detailed comparison of several OOD detection methods, on individual OOD datasets. WideResNet is trained on CIFAR-100 as in-distribution dataset and tested on standard OOD datasets.}
\label{table:results_details_C100}
\end{table}

\subsubsection{Quantitative Results}

In this section, we quantitatively benchmark our approaches against the current state-of-the-art energy based approaches and against several other OOD detection methods including the standard Softmax approach. We benchmark our approach CFL-MLSE where WideResNet is trained as well as tested with semantic energy. 

We showcase the results on 3 evaluation metrics: FPR95, AUROC and AUPR, in Table \ref{table:result}. The table shows the averaged results on 5 OOD test datasets with CIFAR-10 and CIFAR-100 as the in-distribution datasets. Our CFL based Multi-Layer Semantic Energy framework (CFL-MLSE described in Section \ref{section:methodology}) outperforms achieving relative average FPR95 reduction by 67.2\% on CIFAR-10 and 57.4\% on CIFAR-100

Table \ref{table:results_details_C10} and Table \ref{table:results_details_C100} showcases our benchmarked results against state-of-the-art OOD methods on 5 individual testsets. Our approach significantly reduces the relative FPR95 by 18.9\% on CIFAR-10 and by 46.1\% on CIFAR-100 over state-of-the-art energy based model on OOD detection while marginally improving AUROC and AUPR on both the datasets. 






\section{Ablation Analysis}
\label{section:ablation}
\subsection{OOD analysis for classification models}

We conduct ablation studies for further understanding and thorough analysis of our proposed approach. To demonstrate the impact of our methodology on detecting outliers and improving in-distribution accuracy, we provide our ablation study with incremental setups adding our novel features one by one to each as described below. First, we explain our Semantic Energy (SE) setup. Second, we describe an improved framework called Multi Layer Semantic Energy (MLSE). Third, we present CFL based MLSE framework.

1) Semantic Energy (SE) framework \textminus  We study the effectiveness of our approach when the model is trained with ii-loss and our SE loss. Moreover, we also update the mean cluster vector for each mini-batch via EMA during training. This implies that, as the model learns through a semantic energy-bounded objective by assigning lower energies to the in-distribution data, and higher energies to the OOD data, the distribution of matrix $M$ also gets updated for cluster distance calculation.


\begin{table}[t]
\begin{center}
\begin{tabular}{l|l|c|c|c}
\hline
\textbf{In-dataset} & \textbf{Method} & \multicolumn{1}{l|}{\textbf{FPR95}} & \multicolumn{1}{l|}{\textbf{AUROC}} & \multicolumn{1}{l}{\textbf{AUPR}} \\ \hline \hline
CIFAR-10    & SE     & 3.99 & 98.90 & 99.77 \\
            & MLSE  & 3.03 & 99.13  & 99.82 \\
        & CFL-MLSE & \textbf{1.61} & \textbf{99.51} & \textbf{99.89} \\
                    \hline
CIFAR-100      & SE     & 15.72 & 96.07 & 99.04     \\
                    & MLSE  & 13.29 &  96.32 & 99.0 \\ 
                    & CFL-MLSE & \textbf{12.41} & \textbf{97.18}  & \textbf{99.37} \\
                    \hline
\end{tabular}
\end{center}
\caption{Ablation Results of our proposed OOD detection approaches (averaged over 5 OOD datasets). Bold represents superior results. Our proposed approaches: SE (semantic driven energy) MLSE (Multi Layer with SE framework) CFL-MLSE (Cluster Focal Loss with MLSE framework) trained on WideResNet.}
\label{table:ablation}
\end{table}

2) Multi Layer Semantic Energy (MLSE) framework \textminus 
We propose a multiple layer training setting, which we refer to as MLSE. The idea is to explore the effect of considering an aggregated energy score through training multiple layers as an energy based model. To begin with, we incorporate ii-loss~\cite{hassen2020learning} for learning the cluster representation. Next, unlike the SE framework, where only the final layer is leveraged, here multiple layers contribute to energy bounded learning. To be more specific, MLSE and SE have no difference in architecture i.e. both have exactly the same number of model parameters, yet MLSE exceeds SE in performance comprehensively as shown in Table~\ref{table:ablation}.

3) CFL based Multi Layer Semantic Energy (CFL-MLSE) framework \textminus Unlike the SE and MLSE frameworks, in this experiment we introduce our proposed CFL to learn the semantics of class wise cluster centres. To understand the efficacy of CFL, we then train our model end-to-end using the MLSE formulation with ii-loss replaced by our CFL for modeling cluster representation. This training framework involves both our major contribution, and thus we have showcased performance of this model in Tables \ref{table:result}, \ref{table:results_details_C10} and \ref{table:results_details_C100}.

\begin{figure}[t]
\vspace{-5mm}
\begin{center}
   \includegraphics[width=0.99\linewidth]{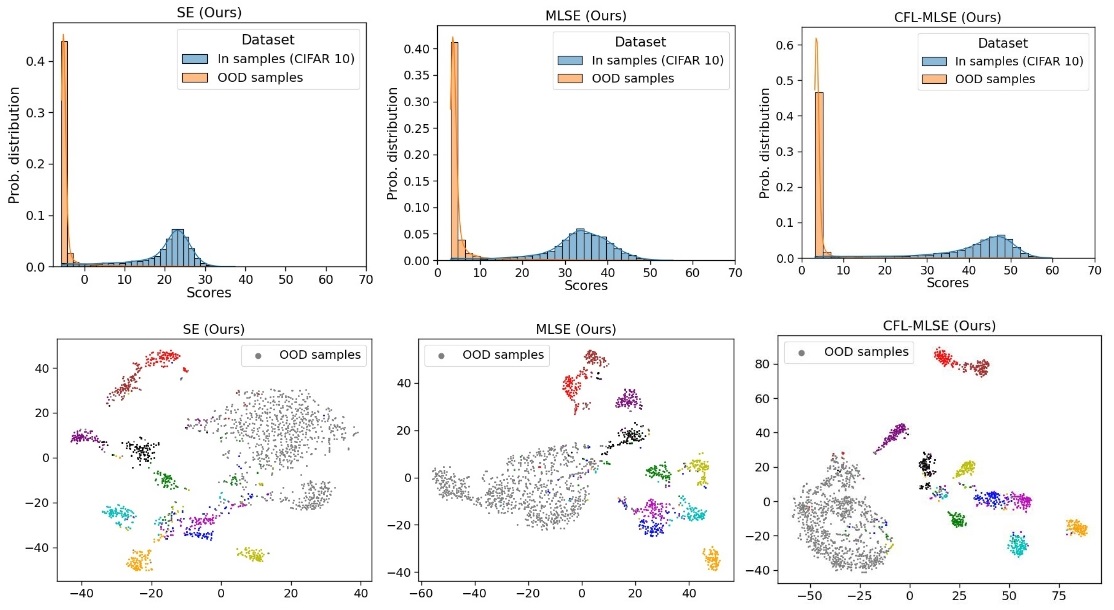}
\end{center}
   \caption{Ablation study of our proposed approaches: SE, MLSE, CFL-MLSE model. It can be observed that CFL(in CFL-MLSE method) further helps to create a greater gap in energy scores between IN-distribution and OOD data, and also forms better clusters.}
\label{fig:CFL_comparison}
\end{figure}

The quantitative results of our ablation are provided in Table \ref{table:ablation}. For all the methods described, the models have the same size and exactly the same number of parameters. It can be observed that all the three proposed methods discussed above perform better than prior art. MLSE has an edge over SE in terms of FPR95 reduction, which can be attributed to the multi-levels involved in it's energy scoring function.  

We further analyze the effect of our proposed CFL training on one of our best performing models. Not only does CFL-MLSE enhances the performance objectively (as shown in Table~\ref{table:ablation}) but CFL efficacy is also evident from the qualitative analysis as illustrated in Figure~\ref{fig:CFL_comparison}. This thorough ablation justifies the superiority of our method.

\subsection{Energy based OOD for Object Detection}

We extend our semantic driven energy based approach from classification to object detector models. We choose to work on a two stage object detector architecture model, to check the effectiveness of our approach on an out-of-distribution dataset. Specifically, we use Faster R-CNN \cite{Ren2015FasterRT} in our experiments.

  In our experiments, we use Pascal VOC 2007 \cite{Pascal_VOC} as IN-distribution dataset. We train the classifier branch with our proposed SE, CFL-SE method. We do not incorporate Multi-Layer based variants of our framework due to feasibility reasons in the Faster R-CNN architecture. For OOD train data, we do not explicitly provide out-of-distribution labels as openset dataset. Instead, we leverage the negative class labels obtained from the proposal target layer \cite{Ren2015FasterRT} during the training and use them as out-of-distribution samples. Without disturbing the regression branch, we train only the classification head with our energy bounded learning approach for 50k iterations with a learning rate of $10^{-4}$ and no weight updates for rest of the network. We keep exactly the same setting to train for vanilla energy \cite{liu2020energy} setup as well.

\begin{table}[b]
\begin{center}
\begin{tabular}{ l | c | c | c | c | } 
  \hline
  \textbf{Testset} & \textbf{Experiment} & \textbf{FPR95} & \textbf{AUROC} & \textbf{AUPR} \\
  \hline
   \hline
   & Softmax & 87.42 & 76.30  & 82.12  \\
\textbf{ MS COCO} & Energy \cite{liu2020energy} & 91.63 & 73.82 & 77.27\\
\textbf{(OUT-distribution)} & SE (Ours) & 83.57 & 78.87 & \textbf{82.69}\\
 & CFL-SE(Ours) & \textbf{73.88} & \textbf{79.59} & 79.69 \\
   \hline

\end{tabular}
\end{center}

\caption{Comparison of OOD performance for Object Detection task. F-RCNN is trained and tested on Pascal VOC train-test dataset as IN-distribution dataset. MS COCO testset is used as out-of-distribution dataset for benchmarking. \vspace{-5mm}}
\label{table:object_detection}
\end{table}
  
  For evaluation, we use MS COCO testset \cite{lin2014microsoft}  and remove the images having overlap with VOC2007 classes. We consider classification and overlapping thresholds of 0.5 and 0.3 respectively for predictions. We benchmark the softmax based pretrained F-RCNN model \cite{Ren2015FasterRT}, vanilla energy \cite{liu2020energy} and our proposed semantic driven energy based model. We keep the same evaluation setup across all the models for a fair comparison. We showcase our results in Table \ref{table:object_detection}. It is observed that both our model's performance exceeds the prior methods, with good reduction in FPR95 for out-of-distribution object detection setting, while maintaining similar numbers on the IN-distribution Pascal VOC testset.
  Hence, outperforming state-of-the-art energy \cite{liu2020energy} based OOD model as object detector.

\section{Conclusion}
In this paper, we proposed a novel and effective semantic driven energy based approach for out-of-distribution (OOD) detection. Our method significantly improves OOD detection on prior state-of-the-art methods. Along with separating in-distribution and out-of-distribution samples, our approach preserves class semantics, thereby improving or maintaining in-distribution accuracy and outperforming the current energy-based approaches and other methods in OOD detection. We also introduce novel Cluster Focal Loss that is majorly focused on learning better representation of class wise cluster centres with maximum inter class separation. This work is largely focused on image classification and two stage object detectors. Future work involves exploring the effectiveness of our approach in video understanding like video classification.

{\small
\bibliographystyle{ieee_fullname}
\bibliography{conference_101719}

\begin{thebibliography}{10}\itemsep=-1pt

\bibitem{bendale2016towards}
Abhijit Bendale and Terrance~E Boult.
\newblock Towards open set deep networks.
\newblock In {\em Proceedings of the IEEE conference on computer vision and
  pattern recognition}, pages 1563--1572, 2016.

\bibitem{cimpoi2014describing}
Mircea Cimpoi, Subhransu Maji, Iasonas Kokkinos, Sammy Mohamed, and Andrea
  Vedaldi.
\newblock Describing textures in the wild.
\newblock In {\em Proceedings of the IEEE Conference on Computer Vision and
  Pattern Recognition}, pages 3606--3613, 2014.

\bibitem{deng2020residual}
Yuntian Deng, Anton Bakhtin, Myle Ott, Arthur Szlam, and Marc'Aurelio Ranzato.
\newblock Residual energy-based models for text generation.
\newblock {\em arXiv preprint arXiv:2004.11714}, 2020.

\bibitem{dhamija2018reducing}
Akshay~Raj Dhamija, Manuel G{\"u}nther, and Terrance~E Boult.
\newblock Reducing network agnostophobia.
\newblock {\em arXiv preprint arXiv:1811.04110}, 2018.

\bibitem{ditria2020opengan}
Luke Ditria, Benjamin~J Meyer, and Tom Drummond.
\newblock Opengan: Open set generative adversarial networks.
\newblock In {\em Proceedings of the Asian Conference on Computer Vision},
  2020.

\bibitem{du2019implicit}
Yilun Du and Igor Mordatch.
\newblock Implicit generation and generalization in energy-based models.
\newblock {\em CoRR}, abs/1903.08689, 2019.

\bibitem{Pascal_VOC}
Mark Everingham, Luc Gool, Christopher~K. Williams, John Winn, and Andrew
  Zisserman.
\newblock The pascal visual object classes (voc) challenge.
\newblock {\em Int. J. Comput. Vision}, 88(2):303–338, June 2010.

\bibitem{grathwohl2019your}
Will Grathwohl, Kuan-Chieh Wang, J{\"o}rn-Henrik Jacobsen, David Duvenaud,
  Mohammad Norouzi, and Kevin Swersky.
\newblock Your classifier is secretly an energy based model and you should
  treat it like one.
\newblock {\em arXiv preprint arXiv:1912.03263}, 2019.

\bibitem{gustafsson2020energy}
Fredrik~K Gustafsson, Martin Danelljan, Goutam Bhat, and Thomas~B Sch{\"o}n.
\newblock Energy-based models for deep probabilistic regression.
\newblock In {\em European Conference on Computer Vision}, pages 325--343.
  Springer, 2020.

\bibitem{haarnoja2017reinforcement}
Tuomas Haarnoja, Haoran Tang, Pieter Abbeel, and Sergey Levine.
\newblock Reinforcement learning with deep energy-based policies.
\newblock In {\em International Conference on Machine Learning}, pages
  1352--1361. PMLR, 2017.

\bibitem{hassen2020learning}
Mehadi Hassen and Philip~K Chan.
\newblock Learning a neural-network-based representation for open set
  recognition.
\newblock In {\em Proceedings of the 2020 SIAM International Conference on Data
  Mining}, pages 154--162. SIAM, 2020.

\bibitem{hendrycks2018deep}
Dan Hendrycks, Mantas Mazeika, and Thomas Dietterich.
\newblock Deep anomaly detection with outlier exposure.
\newblock In {\em International Conference on Learning Representations}, 2019.

\bibitem{hendrycks2019using}
Dan Hendrycks, Mantas Mazeika, Saurav Kadavath, and Dawn Song.
\newblock Using self-supervised learning can improve model robustness and
  uncertainty.
\newblock {\em arXiv preprint arXiv:1906.12340}, 2019.

\bibitem{hsu2020generalized}
Yen-Chang Hsu, Yilin Shen, Hongxia Jin, and Zsolt Kira.
\newblock Generalized odin: Detecting out-of-distribution image without
  learning from out-of-distribution data.
\newblock In {\em Proceedings of the IEEE/CVF Conference on Computer Vision and
  Pattern Recognition}, pages 10951--10960, 2020.

\bibitem{kong2021opengan}
Shu Kong and Deva Ramanan.
\newblock Opengan: Open-set recognition via open data generation.
\newblock {\em arXiv preprint arXiv:2104.02939}, 2021.

\bibitem{lecun2006tutorial}
Yann LeCun, Sumit Chopra, Raia Hadsell, M Ranzato, and F Huang.
\newblock A tutorial on energy-based learning.
\newblock {\em Predicting structured data}, 1(0), 2006.

\bibitem{lee2017training}
Kimin Lee, Honglak Lee, Kibok Lee, and Jinwoo Shin.
\newblock Training confidence-calibrated classifiers for detecting
  out-of-distribution samples.
\newblock {\em arXiv preprint arXiv:1711.09325}, 2017.

\bibitem{Lee2018ASU}
Kimin Lee, Kibok Lee, Honglak Lee, and Jinwoo Shin.
\newblock A simple unified framework for detecting out-of-distribution samples
  and adversarial attacks.
\newblock In {\em NeurIPS}, 2018.

\bibitem{liang2018enhancing}
Shiyu Liang, Yixuan Li, and R. Srikant.
\newblock Enhancing the reliability of out-of-distribution image detection in
  neural networks.
\newblock In {\em International Conference on Learning Representations}, 2018.

\bibitem{Lin2017FocalLF}
Tsung-Yi Lin, Priya Goyal, Ross~B. Girshick, Kaiming He, and Piotr Doll{\'a}r.
\newblock Focal loss for dense object detection.
\newblock {\em 2017 IEEE International Conference on Computer Vision (ICCV)},
  pages 2999--3007, 2017.

\bibitem{lin2014microsoft}
Tsung-Yi Lin, Michael Maire, Serge Belongie, James Hays, Pietro Perona, Deva
  Ramanan, Piotr Doll{\'a}r, and C~Lawrence Zitnick.
\newblock Microsoft coco: Common objects in context.
\newblock In {\em European conference on computer vision}, pages 740--755.
  Springer, 2014.

\bibitem{liu2020energy}
Weitang Liu, Xiaoyun Wang, John Owens, and Yixuan Li.
\newblock Energy-based out-of-distribution detection.
\newblock {\em Advances in Neural Information Processing Systems}, 2020.

\bibitem{maaten2008visualizing}
Laurens van~der Maaten and Geoffrey Hinton.
\newblock Visualizing data using t-sne.
\newblock {\em Journal of machine learning research}, 9(Nov):2579--2605, 2008.

\bibitem{mikolov2013distributed}
Tomas Mikolov, Ilya Sutskever, Kai Chen, Greg~S Corrado, and Jeff Dean.
\newblock Distributed representations of words and phrases and their
  compositionality.
\newblock In {\em Advances in neural information processing systems}, pages
  3111--3119, 2013.

\bibitem{mohseni2020self}
Sina Mohseni, Mandar Pitale, JBS Yadawa, and Zhangyang Wang.
\newblock Self-supervised learning for generalizable out-of-distribution
  detection.
\newblock In {\em Proceedings of the AAAI Conference on Artificial
  Intelligence}, volume~34, pages 5216--5223, 2020.

\bibitem{neal2018open}
Lawrence Neal, Matthew Olson, Xiaoli Fern, Weng-Keen Wong, and Fuxin Li.
\newblock Open set learning with counterfactual images.
\newblock In {\em Proceedings of the European Conference on Computer Vision
  (ECCV)}, pages 613--628, 2018.

\bibitem{Netzer2011ReadingDI}
Yuval Netzer, Tao Wang, Adam Coates, Alessandro Bissacco, Bo Wu, and Andrew~Y.
  Ng.
\newblock Reading digits in natural images with unsupervised feature learning.
\newblock In {\em NIPS Workshop on Deep Learning and Unsupervised Feature
  Learning 2011}, 2011.

\bibitem{ngiam2011learning}
Jiquan Ngiam, Zhenghao Chen, Pang~Wei Koh, and Andrew~Y Ng.
\newblock Learning deep energy models.
\newblock In {\em ICML}, 2011.

\bibitem{NEURIPS2019_9015}
A.~et~al. Paszke.
\newblock Pytorch: An imperative style, high-performance deep learning library.
\newblock In {\em Advances in Neural Information Processing Systems 32}, pages
  8024--8035. Curran Associates, Inc., 2019.

\bibitem{Ren2015FasterRT}
Shaoqing Ren, Kaiming He, Ross~B. Girshick, and J. Sun.
\newblock Faster r-cnn: Towards real-time object detection with region proposal
  networks.
\newblock {\em IEEE Transactions on Pattern Analysis and Machine Intelligence},
  39:1137--1149, 2015.

\bibitem{UMAP}
Tim Sainburg, Leland McInnes, and Timothy~Q. Gentner.
\newblock Parametric umap: learning embeddings with deep neural networks for
  representation and semi-supervised learning.
\newblock {\em ArXiv e-prints}, 2020.

\bibitem{song2020sliced}
Yang Song, Sahaj Garg, Jiaxin Shi, and Stefano Ermon.
\newblock Sliced score matching: A scalable approach to density and score
  estimation.
\newblock In {\em Uncertainty in Artificial Intelligence}, pages 574--584.
  PMLR, 2020.

\bibitem{song2021train}
Yang Song and Diederik~P Kingma.
\newblock How to train your energy-based models.
\newblock {\em arXiv preprint arXiv:2101.03288}, 2021.

\bibitem{wenliang2019learning}
Li Wenliang, Dougal Sutherland, Heiko Strathmann, and Arthur Gretton.
\newblock Learning deep kernels for exponential family densities.
\newblock In {\em International Conference on Machine Learning}, pages
  6737--6746. PMLR, 2019.

\bibitem{xie2016theory}
Jianwen Xie, Yang Lu, Song-Chun Zhu, and Yingnian Wu.
\newblock A theory of generative convnet.
\newblock In {\em International Conference on Machine Learning}, pages
  2635--2644. PMLR, 2016.

\bibitem{Xu2015TurkerGazeCS}
Pingmei Xu, Krista~A. Ehinger, Yinda Zhang, A. Finkelstein, Sanjeev~R.
  Kulkarni, and J. Xiao.
\newblock Turkergaze: Crowdsourcing saliency with webcam based eye tracking.
\newblock {\em ArXiv}, abs/1504.06755, 2015.

\bibitem{yu2015lsun}
Fisher Yu, Ari Seff, Yinda Zhang, Shuran Song, Thomas Funkhouser, and Jianxiong
  Xiao.
\newblock Lsun: Construction of a large-scale image dataset using deep learning
  with humans in the loop.
\newblock {\em arXiv preprint arXiv:1506.03365}, 2015.

\bibitem{zagoruyko2016wide}
Sergey Zagoruyko and Nikos Komodakis.
\newblock Wide residual networks.
\newblock {\em arXiv preprint arXiv:1605.07146}, 2016.

\bibitem{Zhou2018PlacesA1}
Bolei Zhou, {\`A}. Lapedriza, A. Khosla, A. Oliva, and A. Torralba.
\newblock Places: A 10 million image database for scene recognition.
\newblock {\em IEEE Transactions on Pattern Analysis and Machine Intelligence},
  40:1452--1464, 2018.

\end{thebibliography}
}

\end{document}